\documentclass[10pt]{wlscirep}

\usepackage[utf8]{inputenc}
\usepackage[T1]{fontenc}
\usepackage{graphicx}
\usepackage{booktabs}
\usepackage{multirow}
\usepackage{array}
\usepackage{amsmath}
\usepackage{caption}
\usepackage{algorithm}
\usepackage{algorithmic}
\usepackage{setspace}
\usepackage{placeins}
\usepackage{url}
\usepackage{float}
\usepackage{marvosym}
\graphicspath{{图片/}{figures/}{../figures/}}
\begin{document}
\doublespacing

\title{A Self-Evolving Agent for Longitudinal Personal Health Management}

\author[1,+]{Haoran Li}
\author[1,+]{Jiebi Deng}
\author[6]{Tong Jin}
\author[2]{Jinghong Han}
\author[2]{Yuxin Wang}
\author[7]{Zexin Wang}
\author[8]{Jinyu Sun}
\author[5]{Qingyi Si}
\author[1]{Weikang Gong}
\author[1]{Xiahai Zhuang}
\author[2,4,\Letter]{Jia You}
\author[2,3,\Letter]{Wei Cheng}
\author[2,\Letter]{Jianfeng Feng}
\author[1,\Letter]{Hongcheng Guo}

\affil[1]{School of Data Science, Fudan University, Shanghai, China}
\affil[2]{Institute of Science and Technology for Brain-Inspired Intelligence, Fudan University, Shanghai, China}
\affil[3]{Department of Neurology, Huashan Hospital, Fudan University, Shanghai, China}
\affil[4]{Key Laboratory of Computational Neuroscience and Brain-Inspired Intelligence, Fudan University, Ministry of Education, Shanghai, China}
\affil[5]{JD.com, Inc., Beijing, China}
\affil[6]{School of Life Sciences, Beijing University of Chinese Medicine, Beijing, China}
\affil[7]{School of Computer Science and Technology, Huazhong University of Science and Technology, Wuhan, China}
\affil[8]{Nanjing Medical University, Nanjing, China}
\affil[+]{Contributes equally}
\affil[\Letter]{Corresponding author}

\begin{abstract}
Medical large language models and agents show strong single-encounter question answering and evidence interpretation but struggle to accumulate and govern experience across encounters. We present HealthJev, an open-source agent separating shared safety rules and medical knowledge from a private longitudinal memory and distilling lessons from completed cases. To filter out erroneous lessons before they affect future behaviour, we self-train JevGate, a lightweight System-One model that decides which lessons enter memory; each decision takes a median of 86 milliseconds and about 1{,}000 input tokens, sustaining experience accumulation at negligible run-time overhead. On a longitudinal benchmark of 90-day simulated daily interactions for 20 users built from real MIMIC-IV trajectories, HealthJev outperformed a memory-free ReAct agent and three memory-augmented frameworks, with all paired comparisons significant. Across six tasks spanning clinical records, medical imaging, genomics and clinical text, HealthJev improved every task with a mean gain of 13.3 percentage points over the base condition, surpassing all baselines. The frozen library transferred to six further backbones with macro-average gains of 6.8 to 15.1 percentage points, and one task's library brought gains on other tasks. These results indicate that governed, self-evolving memory supports longitudinal personal health management beyond single encounters. HealthJev is publicly available at \url{https://github.com/HC-Guo/Self-Evolving_HealthJev}.

\end{abstract}

\maketitle

\section*{Main}

Personal health management is shaped by what happens after advice is given, not only by the advice itself. Meal plans must accommodate delayed dinners, changing work schedules and preferences that become clear only through use. Reminders become less effective when routines shift, while symptoms and measurements acquire meaning through their trajectory rather than any single reading. Wearable sensors, digital biomarkers and mobile self-management programmes have made these patterns easier to observe, but support still loses value when it cannot adapt to daily feedback and sustained use \cite{eysenbach2005attrition,strain2020wearable,coravos2019digitalbiomarkers,goldsack2020v3,moschonis2023digitaldiabetes}. A personal health agent must therefore allow repeated encounters to change later support for the same person, rather than treating every request as the first.

Existing systems address this need only in part. Foundation models and large language models now support medical question answering, diagnostic reasoning and conversational diagnosis within a single encounter \cite{rajpurkar2022healthcareai,thirunavukarasu2023llmmedicine,singhal2023clinicalknowledge,singhal2025expert,mcduff2025differential,tu2025conversationaldiagnostic}. Symptom checkers and clinical decision-support systems structure assessment and triage \cite{fraser2023ada,ribolisasco2023symptom,fda2026cds}, and newer personal-health systems translate sleep, activity and wearable streams into user-facing guidance \cite{khasentino2025phllm,merrill2026phia}. Medical-agent research has also begun to formalize action spaces and workflow evaluation \cite{wang2025medicalagents}. In parallel, memory-augmented agents retain, consolidate and retrieve prior experience to alter subsequent behaviour \cite{packer2023memgpt,zhong2024memorybank,chhikara2025mem0,li2025memp}, and self-improving agents distil reusable skills and lessons from their own trajectories \cite{shinn2023reflexion,wang2023voyager,zhao2024expel}. Recent work further observes that many of these memory decisions require semantic judgement rather than text generation, and delegates them to lightweight, non-generative controllers \cite{jiang2026jevmem}. These lines of work have developed largely in isolation: encounter-level medical AI rarely carries state across visits, general-purpose agent memory is primarily evaluated on open-domain dialogues and tasks rather than clinical trajectories, and neither fully addresses the persistence and disclosure decisions that personal health data demand.

Personal health makes this gap a concrete design problem: selective memory updates, not recall alone. One encounter may contain a stable allergy, a transient symptom, a failed recommendation, an improved meal-planning rule, an outdated measurement and a sensitive identifier, and these items should not share the same persistence or disclosure policy. Supplying the full history at every turn can aid recall, but it also increases context cost, irrelevant-history noise and potential exposure of identifiable health data \cite{price2019privacy,rieke2020federated,kaissis2020secure}; withholding history instead reduces that exposure but discards the continuity that longitudinal care requires. Both extremes have limitations at scale. What is needed is a memory that decides, encounter by encounter, what should be kept, revised or withheld, and candidate lessons should be checked before admission rather than accepted by default, because a lesson distilled from a single case can be incorrect. Once an erroneous lesson enters memory, it may be retrieved into later reasoning and lead to incorrect answers, reducing the quality of subsequent support. Because these checks recur at every encounter, their cost is itself part of the design problem: a large language model can judge each candidate carefully, but delegating thousands of routine judgements to it can be slow and expensive, while fixed rules are inexpensive but brittle.

We developed HealthJev to make these decisions explicit. The agent separates shared medical knowledge and safety constraints from a private longitudinal memory of profile facts, reusable procedures and episodic traces, and updates this memory as each case unfolds. We evaluated the design on a longitudinal benchmark constructed from the real disease trajectories of 20 MIMIC-IV patients, with 90 days of simulated daily interaction per user; to our knowledge, this is the first agent benchmark for longitudinal personal health management built from real clinical trajectories. HealthJev outperformed a memory-free ReAct agent and three widely used memory-augmented frameworks mounted on an identical scaffold in every paired comparison against these baselines. Beyond a single benchmark, we asked whether the experience an agent accumulates is itself reusable. On six heterogeneous health-AI tasks, the agent distils a candidate lesson after each completed case, and JevGate, a lightweight, self-trained System-One decision model, filters erroneous candidates out of subsequent retrieval. JevGate makes admission decisions with a median latency of 86 milliseconds and a 90th-percentile latency of 184 ms, supporting continued experience accumulation and governance at low decision latency. With the experience libraries accumulated this way, HealthJev improved performance on every task with a mean gain of 13.3 percentage points over the base condition. When frozen and mounted on six further language-model backbones, these libraries improved the macro-average metric on all six backbones, and lessons accumulated on one task also aided other tasks that had not been used to build the source library.

\section*{Results}

\subsection*{A self-evolving memory loop for personal health}

HealthJev evaluates the information from each completed episode to determine whether it should be retained for subsequent support. A stable allergy, transient symptom, unsuccessful recommendation, planning rule and sensitive identifier may occur in the same exchange, but require different retention and update policies. Post-episode induction determines how each item will be used, rather than appending the full dialogue to the next prompt.

\begin{figure}[p]
\centering
\includegraphics[width=\linewidth,height=0.9\textheight,keepaspectratio]{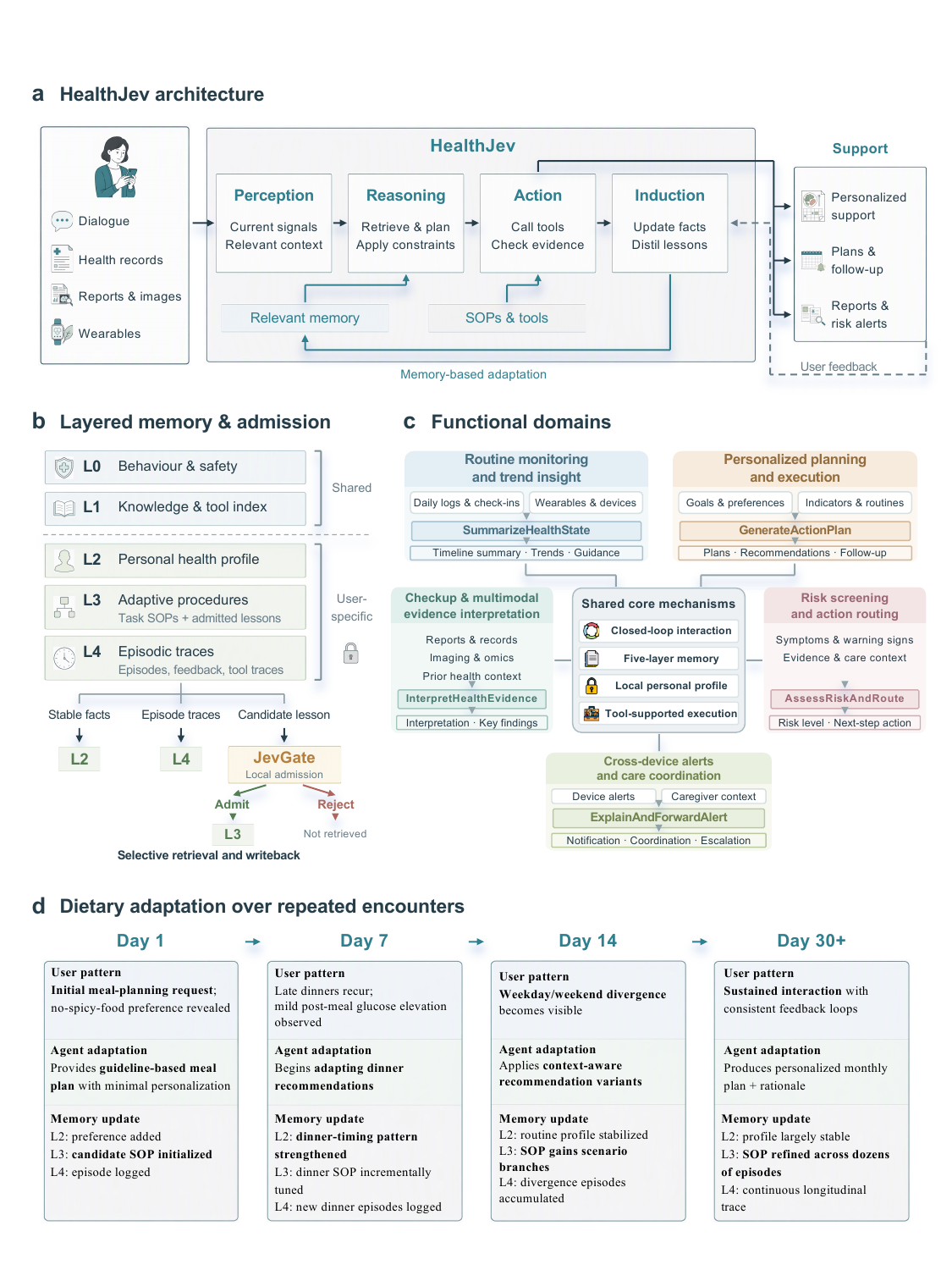}
\caption{\textbf{HealthJev architecture, layered memory and longitudinal adaptation.} \textbf{a,} Closed loop of perception, reasoning, action and post-episode induction. \textbf{b,} Shared behaviour and safety rules (L0) and knowledge and tool index (L1), with user-specific profiles (L2), reusable procedures (L3) and episodic traces (L4); JevGate screens candidate lessons for L3 admission. \textbf{c,} Five functional domains sharing the same interaction, memory and tool mechanisms. \textbf{d,} Simulated dietary adaptation from day 1 to day 30+, illustrating recommendation and memory updates, not measured clinical outcomes.}
\label{fig:architecture}
\end{figure}

HealthJev implements this process as a closed loop of perception, reasoning, action and post-episode induction (Fig.~\ref{fig:architecture}a). Perception assembles the current request, available health signals and relevant prior context. Reasoning combines task-relevant memory with shared medical knowledge to plan under user-specific constraints and safety rules. Action produces the response or executes the task workflow. After the episode, induction determines whether the information should update profile facts, revise a reusable procedure, be retained as an episodic trace or be excluded from future use. JevGate screens candidate lessons for admission to the reusable experience library. This lightweight, self-trained System-One decision model has a median admission-decision latency of 86 ms and a 90th-percentile latency of 184 ms. Self-evolution refers here to the influence of completed encounters on subsequent interactions through updates to memory and reusable procedures.

Shared knowledge and governance rules occupy L0--L1, whereas user-specific state occupies L2--L4 (Fig.~\ref{fig:architecture}b). L2 holds durable profile facts, L3 reusable procedures and L4 episodic traces. This division supports task-specific retrieval without supplying the complete longitudinal record. Sensitive details can remain local, be minimized or be excluded from writeback. The closed loop and layered memory separate shared knowledge and governance rules from user-specific state, allowing subsequent interactions to use selectively retained and updated information rather than the full dialogue history.

The architecture supports five representative personal-health workflows (Fig.~\ref{fig:architecture}c). A simulated dietary-management case illustrates how profile facts, task procedures and episodic traces evolve across repeated encounters (Fig.~\ref{fig:architecture}d).

\subsection*{A longitudinal benchmark tests personal health management}

\begin{figure}[!htbp]
\centering
\includegraphics[width=\linewidth,height=0.85\textheight,keepaspectratio]{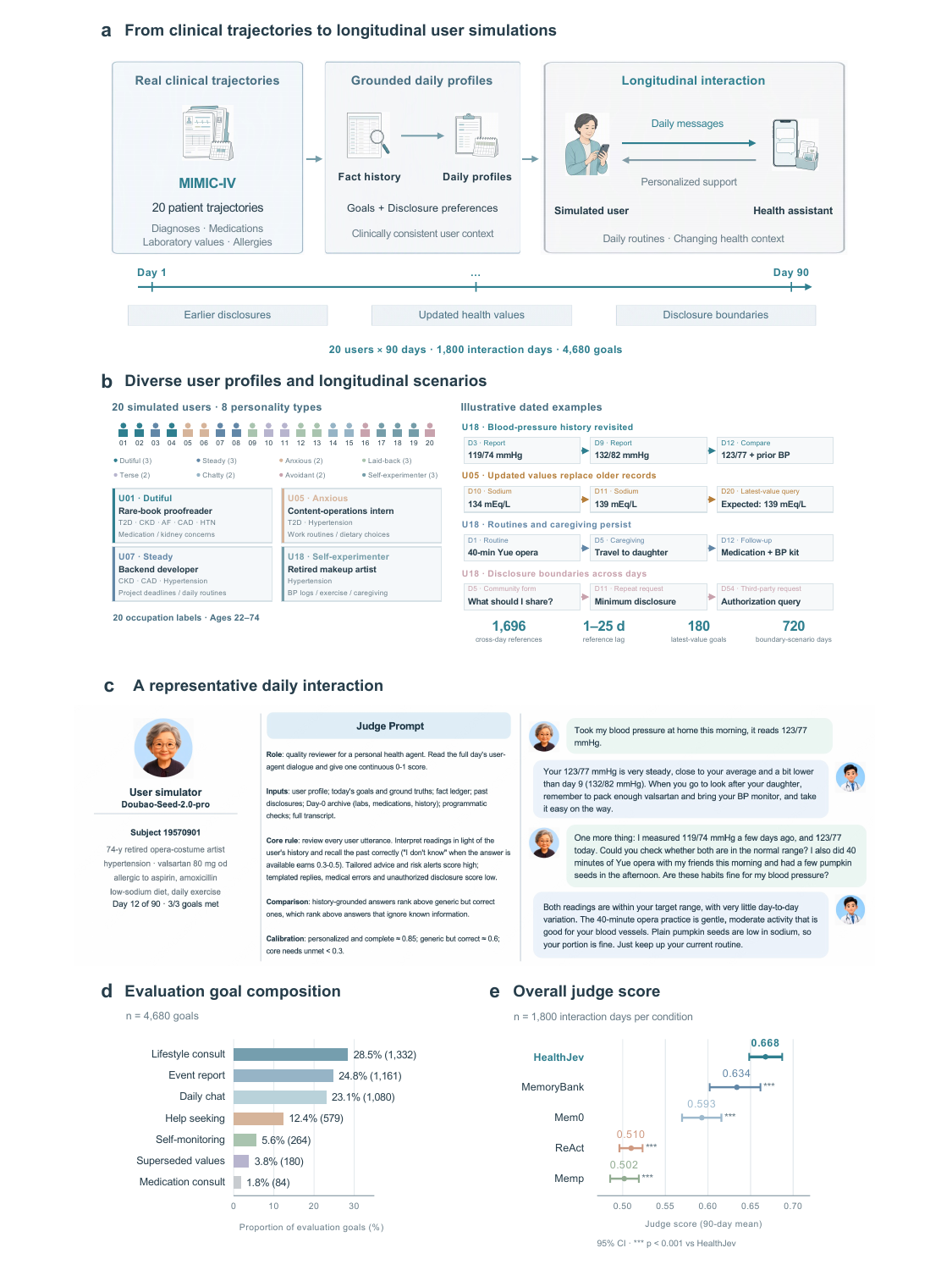}
\caption{\textbf{A record-grounded longitudinal benchmark for personal health management.} \textbf{a,} MIMIC-IV trajectories underpin fact histories and daily profiles for 20 simulated users over 90 days. \textbf{b,} Anonymous user profiles and examples of historical references, value updates, recurring routines and disclosure boundaries; user glyph colours identify eight persona categories. \textbf{c,} Representative user--HealthJev dialogue and blinded GLM-5.2 judging prompt. \textbf{d,} Composition of 4,680 goals across seven types; bars show percentages and counts. \textbf{e,} Dots and horizontal error bars show mean judge scores over 1,800 interaction days per condition and user-level cluster-bootstrap 95\% confidence intervals (10,000 resamples). $***$: $p<0.001$, two-sided paired Wilcoxon signed-rank tests against HealthJev.}
\label{fig:benchmark}
\end{figure}

To evaluate longitudinal personal-health support, we constructed a benchmark from the disease trajectories of 20 real MIMIC-IV patients (Fig.~\ref{fig:benchmark}a). Diagnoses, medications and laboratory values from each patient record were used to establish a per-user fact history. Each user underwent 90 days of simulated daily interaction, yielding 1,800 scored interaction days. An LLM user simulator set two to three goals per day and reported record-grounded events and measurements in colloquial first-person messages. The fact history recorded every disclosed value to check the interaction content and constrain the factual basis of the evaluation. The benchmark contained 4,680 goals across seven types: lifestyle consults accounted for 28.5\%, event reports for 24.8\%, daily chat for 23.1\%, help seeking for 12.4\%, self-monitoring for 5.6\%, superseded values for 3.8\% and medication consults for 1.8\% (Fig.~\ref{fig:benchmark}d). The evaluation also included memory and governance probes. Longitudinal recall probes referenced information disclosed 1 to 25 days earlier and occurred on 85.8\% of days; most were cued by the user's words, whereas the remainder required recall without an explicit prompt. A total of 189 registered governance probes, each audited on one or two pre-specified days, yielded 294 governance checks assessing four behaviours: retaining key medical information, not storing sensitive everyday details, correcting misreported values and restricting third-party disclosures to authorized information.

We compared five conditions under the same interaction protocol using Doubao-Seed-2.0-pro as the backbone: HealthJev, a memory-free ReAct agent and three memory-augmented frameworks, MemoryBank, Mem0 and Memp. After each day, the automated judge GLM-5.2 scored the day's dialogue from 0 to 1 against the user's goals and record ground truth. The rubric covered six dimensions: long-term recall, temporal validity of previously reported values, numerical accuracy including correction of misreported values, response to critical readings, privacy boundaries over what is stored and disclosed, and overall fulfilment of the user's needs (Fig.~\ref{fig:benchmark}c). The analysis covered all 1,800 planned interaction days per condition.

HealthJev achieved the highest mean judge score, 0.668, compared with 0.634 for MemoryBank, 0.593 for Mem0, 0.510 for ReAct and 0.502 for Memp (Fig.~\ref{fig:benchmark}e). Paired two-sided Wilcoxon signed-rank tests across the 1,800 interaction days showed significant differences in all four comparisons between HealthJev and the baselines, with the largest $p$ approximately $1.1\times10^{-4}$. At the user level, HealthJev achieved a higher mean daily score than MemoryBank, the strongest baseline, for 15 of the 20 patients.

\subsection*{Performance and experience transfer across biomedical tasks}

\begin{figure}[!htbp]
\centering
\includegraphics[width=0.9\linewidth]{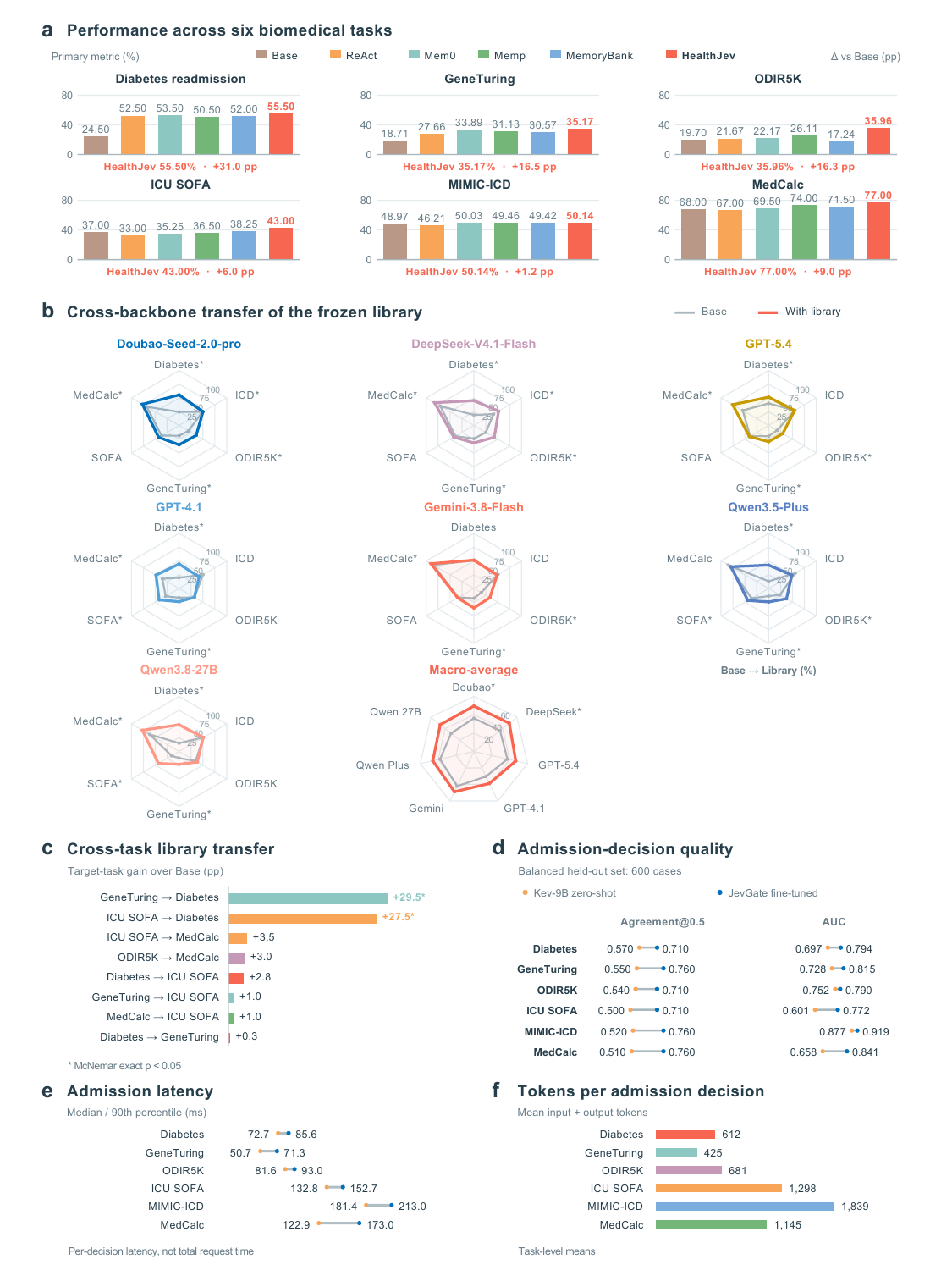}
\caption{\textbf{Biomedical performance, experience transfer and admission governance.} \textbf{a,} Six-task primary metrics (\%) and HealthJev gains over Base (pp). \textbf{b,} Frozen-library transfer across backbones; grey/coloured polygons: Base/Library, including macro-averages. \textbf{c,} Cross-task gains (pp), coloured by donor. \textbf{d,} Agreement@0.5 and AUC for zero-shot Kev-9B (orange) versus fine-tuned JevGate (blue) on 600 balanced held-out cases. \textbf{e,} Per-decision median (orange) and 90th-percentile (blue) latency (ms). \textbf{f,} Mean input plus output tokens per decision. Asterisks: positive gains with two-sided $p<0.05$ in \textbf{b,c}; exact McNemar tests on case scores $>0$ for task comparisons and exact Wilcoxon tests over six tasks for macro-averages.}
\label{fig:biomedical}
\end{figure}

Personal health management involves imaging, laboratory trends, physiological
signals, electronic records and molecular information. We evaluated HealthJev
on six biomedical tasks spanning these evidence types: diabetes readmission
prediction from electronic records, ICU SOFA estimation from physiological
signals, MIMIC-ICD discharge-summary coding, MedCalc medical
calculation, ODIR5K fundus-image classification and GeneTuring genomic
question answering (Fig.~\ref{fig:biomedical}a).

On each task, HealthJev accumulated experience on a split disjoint from the
test set. After each completed case, a candidate lesson was distilled and
screened by JevGate; admitted lessons were available to subsequent cases
through retrieval. All tasks used the same framework, driven by
Doubao-Seed-2.0-pro, with task-specific workflows.

HealthJev achieved a higher task-specific primary metric than the base
condition and all memory-augmented baselines on each of the six tasks
(Fig.~\ref{fig:biomedical}a). The mean absolute gain over the base condition
was 13.3 percentage points. The largest gains were observed for diabetes
readmission, GeneTuring and ODIR5K fundus, with increases of 31.0, 16.5 and
16.3 percentage points, respectively. All three gains were significant by a two-sided McNemar
exact test, with $p<0.001$. MedCalc improved by 9.0 percentage points,
with $p=0.022$. For MIMIC-ICD, the proportion of cases receiving a non-zero
score increased significantly, with $p=0.008$, and mean set F1 increased by
1.2 percentage points. ICU SOFA improved by 6.0 percentage points, but the
difference did not reach statistical significance, with $p=0.13$.

We then froze the experience library accumulated with Doubao-Seed-2.0-pro and
mounted it, without retraining, on six further language-model backbones:
DeepSeek-V4.1-Flash, GPT-5.4, GPT-4.1, Gemini-3.8-Flash, Qwen3.5-Plus and
Qwen3.8-27B. The macro-average metric improved on all six backbones, with
gains of 6.8 to 15.1 percentage points (Fig.~\ref{fig:biomedical}b),
supporting the transferability of the frozen library across backbone models.

\begin{figure}[!htbp]
\centering
\includegraphics[width=\linewidth,height=0.8\textheight,keepaspectratio]{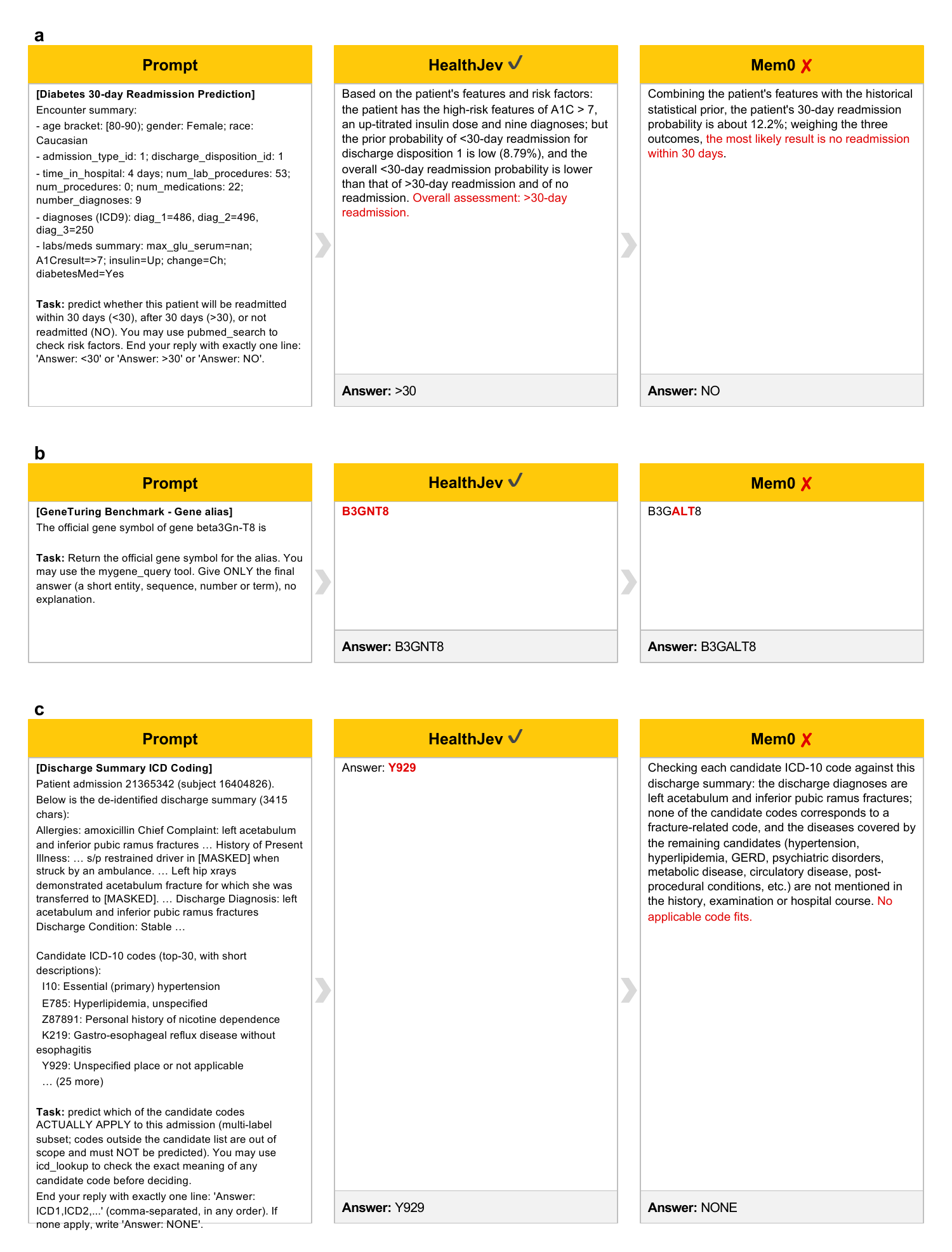}
\caption{\textbf{Representative cases across three biomedical evidence types.} Each row shows one case: the shared prompt (left), the HealthJev reply (middle, tick) and the Mem0 reply (right, cross). Red text marks the decisive evidence in the prompt or the answer, and the strip at the bottom of each reply gives the final answer; HealthJev matched the reference on all three cases, whereas Mem0 did not.
\textbf{a,} Diabetes readmission. Both agents saw the same statistical prior, which placed the probability of readmission within 30 days at about 12\%. Mem0 followed the prior and predicted no readmission; HealthJev weighed poor glycaemic control indicated by A1C $>$ 7, up-titrated insulin and nine diagnoses against the prior and assigned the $>$30-day class.
\textbf{b,} GeneTuring gene-alias question. HealthJev returned the official gene symbol B3GNT8, whereas Mem0 returned the similar symbol B3GALT8.
\textbf{c,} Discharge-summary coding. Fracture codes were outside the candidate set, leaving Y92.9 as the only applicable candidate for the event described, in which the patient was struck by an ambulance. Mem0 concluded that no candidate applied, whereas HealthJev selected this code.}
\label{fig:cases}
\end{figure}

Because every candidate lesson is screened by JevGate, we further evaluated
admission-decision quality. JevGate is built on the Kev-9B decision model.
On a balanced held-out set of 600 admission cases, with 50 reference-admitted
and 50 reference-rejected candidates per task, zero-shot Kev-9B performed near
chance, with a pooled agreement of 0.532. After fine-tuning on labelled
admission data, pooled agreement increased to 0.735, with task-level values of
0.710--0.760, and the AUC pooled across all 600 cases increased from 0.733 to
0.810 (Fig.~\ref{fig:biomedical}d).

We also measured the run-time overhead of admission decisions. Across the
2,983 admission decisions in the six-task evaluation, median decision
latency was 86 ms, and the 90th-percentile latency was 184 ms
(Fig.~\ref{fig:biomedical}e).
Each decision consumed on average about 1,000 input and 22 output tokens
(Fig.~\ref{fig:biomedical}f), and 81.8\% of candidate lessons were admitted.

The frozen library also improved performance in cross-task evaluation.
Transfer across eight donor--target task pairs yielded gains of 0.3 to
29.5 percentage points over the target task's base condition. The largest
gains came from transferring the GeneTuring and ICU SOFA libraries to
diabetes readmission, with gains of 29.5 and 27.5 percentage points,
respectively. Both gains were significant by a two-sided McNemar exact
test, with $p<0.05$ (Fig.~\ref{fig:biomedical}c). Together with the cross-backbone results,
these findings support the transferability of the experience library
across tasks and models.

Three representative cases illustrate differences between the answers of
HealthJev and Mem0, the strongest baseline on the corresponding task
(Fig.~\ref{fig:cases}). HealthJev matched the reference answer in all three
cases, whereas Mem0 did not.

In the diabetes-readmission case, both agents received the same statistical
prior, which placed the probability of readmission within 30 days at about
12\%. Mem0 relied on this prior and predicted no readmission; HealthJev's
answer incorporated poor glycaemic control indicated by A1C $>$ 7,
up-titrated insulin and nine diagnoses and assigned the $>$30-day class.

In the gene-alias question, HealthJev returned the official gene symbol
B3GNT8, whereas Mem0 returned the similar symbol B3GALT8.

In the discharge-summary coding case, fracture codes were outside the
candidate set, and Y92.9 was the only applicable candidate for the event
described. Mem0 concluded that no candidate applied, whereas HealthJev
selected this code. These cases illustrate differences in the handling of
case-specific evidence, gene symbols and candidate-code constraints.

\section*{Discussion}

Longitudinal health support requires continuity while distinguishing different types of information in a person's history. HealthJev assigns profile facts, reusable procedures and episodic traces different update and retrieval rules, while keeping medical knowledge and safety constraints separate from user-specific memory. An admission step determines which lessons from an encounter should be retained. The contribution of this design is explicit control over how prior encounters affect subsequent planning, together with a mechanism for accumulating screened, reusable experience across repeated interactions.

This design connects two lines of work that have largely developed separately. Medical language models and personal-health assistants support question answering, diagnostic reasoning, coaching and wearable-data interpretation \cite{singhal2025expert,mcduff2025differential,khasentino2025phllm,merrill2026phia}. Memory and self-improving agents use prior trajectories to alter subsequent behaviour \cite{packer2023memgpt,shinn2023reflexion,wang2023voyager,zhao2024expel}. Recent work further delegates memory decisions that require semantic judgement but not text generation to lightweight, non-generative controllers \cite{jiang2026jevmem}. HealthJev brings trajectory-based adaptation into a person-level health setting, where retained information may include diagnoses, medication history, routines and family context, and a lesson distilled from one encounter can inform subsequent interactions.

The accumulated experience library transferred across models and tasks. When frozen and mounted on six further language-model backbones, it improved the six-task macro-average metric on all of them by 6.8 to 15.1 percentage points. Cross-task transfer yielded gains of 0.3 to 29.5 percentage points across eight donor--target task pairs, with the largest gains when the GeneTuring and ICU SOFA libraries were transferred to diabetes readmission. These results indicate that the library's utility is not restricted to the model or source task used during experience accumulation. Self-evolution in HealthJev thus involves the formation of screened experience from completed cases and its reuse in subsequent cases, other tasks and different backbone models.

Continued experience accumulation requires a balance between admission-decision quality and run-time overhead. Delegating large numbers of routine judgements to a general-purpose language model increases latency and cost, whereas fixed rules have limited capacity to handle complex semantics. JevGate uses a lightweight, self-trained decision model: on a balanced held-out set of 600 admission cases across six tasks, it increased pooled agreement from 0.532 to 0.735 and pooled AUC from 0.733 to 0.810. Its median decision latency was 86 ms, and each decision consumed on average about 1,000 input and 22 output tokens. It serves as the admission controller during experience accumulation. The three representative cases illustrate differences in the use of statistical priors, identification of gene symbols and selection of candidate codes, emphasizing the need to handle such task-specific information accurately in subsequent reasoning.

The 90-day benchmark evaluates continuity and disclosure within the same patient trajectories. Compared with isolated questions, static records and short conversations, it allows evaluation of whether earlier events affect subsequent support and whether remembered information is disclosed unnecessarily. The user simulator is driven by a fact history grounded in the source record, providing a defined basis for reference facts and privacy probes. Simulation still cannot fully reproduce the ambiguity, missingness and behavioural feedback of prospective use.

Memory governance also provides a point at which privacy-related behaviour can be inspected. Sensitive profile facts, reusable procedures and episode traces can be retrieved or withheld independently, and privacy-sensitive interactions can be excluded from writeback. These mechanisms help reduce unnecessary exposure within the agent loop, but deployment-level privacy still depends on access control, security testing and data governance.

The six-task evaluation examined experience accumulation and use across heterogeneous biomedical evidence. HealthJev improved the primary metric on all six tasks, with statistically significant gains for diabetes readmission, GeneTuring, ODIR5K and MedCalc. The proportion of MIMIC-ICD cases receiving a non-zero score also increased significantly. The frozen library further improved the macro-average metric on six additional backbone models. Under these evaluation conditions, the results support governed experience accumulation and its use in task reasoning rather than zero-shot clinical generalization. The gain on ICU SOFA was 6.0 percentage points but did not reach statistical significance.

The evidence remains limited to offline benchmarks. The 90-day longitudinal interactions were simulated and graded by the automated judge GLM-5.2, without human ratings. Prospective studies should evaluate sustained use, clinician and user judgement, privacy and security, and stability under sparse feedback, contradiction and distribution shift. Such studies require deployment-specific protocols and human oversight under established reporting and evaluation standards \cite{nagendran2020aiworkflow,liu2020consortai,cruzrivera2020spiritai,vasey2022decideai,ong2025governance}.

\bibliography{ref}

\section*{Methods}

\subsection*{Representative functional domains}

The HealthJev architecture was mapped to five representative functional domains (Fig.~\ref{fig:architecture}c). These domains cover routine monitoring and trend analysis from daily logs and wearable signals; personalized planning and execution based on goals, preferences and recent indicators; checkup and multimodal evidence interpretation across reports, imaging and omics-style inputs; risk screening and action routing from heterogeneous signals to urgency levels and next-step recommendations; and cross-device alerting and care coordination for abnormal events. These domains define the intended scope of the architecture.

These domains are representative rather than exhaustive and describe recurring personal-health workflows. Each uses the same underlying architecture: a closed perception--reasoning--action--induction loop, layered memory that separates shared knowledge and governance rules from user-specific state, and tool-supported execution.

\subsection*{HealthJev architecture}

HealthJev treats personal health support as a sequence of episodes in which completed interactions can affect subsequent behaviour through memory updates. The architecture is summarized in Fig.~\ref{fig:architecture}. In each episode, the agent receives the current request, visible health signals, task metadata and retrieved memory. The output may be a conversational response, a structured recommendation, an interpretation of health information or a tool-supported task result.

The interaction loop has four stages. Perception assembles the current request with available health context and relevant prior memory. Reasoning combines this information with shared health knowledge and safety constraints to form a task plan. Action generates the response or executes the workflow, including tool calls when structured evidence is needed. After the episode, induction determines what should be retained, revised or stored as a time-stamped trace. This process updates the information and procedures used in subsequent support. Candidate lessons for the reusable experience library are screened by JevGate, a self-trained admission model; its training is described below.

Memory is separated into five layers (Fig.~\ref{fig:architecture}b). L0 stores behavioural rules, safety boundaries and escalation conditions. L1 stores shared health knowledge, task knowledge and tool-routing information. L2 stores the personal profile, including chronic conditions, allergies, medication-related information, stable preferences, lifestyle constraints and long-term goals. L3 stores reusable procedures for recurring tasks, such as meal planning, indicator tracking, medication follow-up, report interpretation, risk routing and privacy disclosure handling. L4 stores episodic traces, including local context, previous outputs, feedback and tool traces.

Memory writeback is determined by the expected future use of each item. Stable user facts can update L2. Recurring task patterns can revise L3. Local context, transient symptoms and single-episode details can remain in L4. Material that is unnecessary or inappropriate to retain is excluded from long-term memory. L0 and L1 are treated as shared knowledge and governance layers and are not updated by user-level episodes.

The same organization is used to limit unnecessary exposure of sensitive information. HealthJev does not provide the full longitudinal record to every request. Retrieval is restricted to profile facts, procedures or episodic fragments needed for the current task. Sensitive details can remain local, be represented in minimized form or be excluded from writeback. Privacy probes were evaluated with writeback disabled for the probe interaction.

\subsection*{JevGate training}

JevGate is built on the Kev-9B decision model, consisting of a frozen Qwen3.5-9B base model, rank-16 low-rank adaptation (LoRA) and a pointer head. The LoRA weights and pointer head are initialized from the publicly released Kev-9B checkpoint and further fine-tuned on admission data. The base weights remain frozen throughout; only the LoRA weights and pointer-head parameters are updated.

Training records come from real admission decisions made during experience accumulation. Each sample pairs the context of the originating case with the candidate lesson. Let $t_i$ denote the task, $q_i$ the case prompt, $g_i$ the reference answer, $a_i$ the agent's answer, $r_i$ the programmatic score, $e_i$ the error hint, $c_i$ the number of tool calls and $x_i$ the candidate lesson, including its scope and main points. The input state is
\begin{equation}
 s_i = \operatorname{Serialize}(t_i,q_i,g_i,a_i,r_i,e_i,c_i,x_i),
 \label{eq:jev_state}
\end{equation}
where $\operatorname{Serialize}$ organizes these fields into a fixed text format. The case prompt, reference answer, agent answer and error hint are limited to 9{,}500, 1{,}000, 2{,}000 and 200 characters, respectively, and the encoded state-token budget is set to 3{,}072.

The admission label $y_i\in\{0,1\}$ indicates whether a candidate lesson should be retained. Labels are obtained from external-judge records, with the case context, reference answer and programmatic scoring result provided as evidence. A candidate lesson must contain a concrete rule applicable to subsequent cases and be consistent with the recorded problem or error. Cleaning, deduplication and exclusion of held-out cases yielded a merged pool of 13{,}315 records. Positive-label samples for which the previous gate verdict agreed with the reference admission label were downsampled, whereas discordant samples were sampled twice. The resulting training set contained 11{,}290 records, including 7{,}903 positive and 3{,}387 negative labels; these counts include repeated samples.

The pointer head directly scores rejection and admission options rather than generating a verdict as text. Let $h_i^{d}$ denote the hidden state at the decision position and $h_{i,k}^{o}$ the hidden state at the position corresponding to option $k$, where $k=0$ denotes rejection and $k=1$ admission. Two linear projections produce the option scores:
\begin{equation}
 z_{i,k} = \frac{
 (W_K h_{i,k}^{o}+b_K)^{\mathsf T}
 (W_Q h_i^{d}+b_Q)
 }{\sqrt{d_p}}, \qquad d_p=256.
 \label{eq:jev_pointer}
\end{equation}
Here, $W_Q,b_Q$ and $W_K,b_K$ are the projection parameters for the decision and option states, respectively. Both options share the same option projection, and $d_p$ is the pointer dimension. The scores are normalized by softmax:
\begin{equation}
 p_\theta(y=k\mid s_i)
 = \frac{\exp(z_{i,k})}{\exp(z_{i,0})+\exp(z_{i,1})},
 \qquad k\in\{0,1\}.
 \label{eq:jev_probability}
\end{equation}

Let $\theta$ denote the trainable LoRA and pointer-head parameters and $N$ the number of training records after sampling. Training minimizes binary cross-entropy over the admission labels:
\begin{equation}
 \mathcal L(\theta)
 = -\frac{1}{N}\sum_{i=1}^{N}
 \left[
 y_i\log p_\theta(1\mid s_i)
 +(1-y_i)\log p_\theta(0\mid s_i)
 \right].
 \label{eq:jev_loss}
\end{equation}
Each sampled record receives equal loss weight. The programmatic case score is an input feature, not a loss weight.

Optimization uses AdamW with weight decay $0.01$, a peak learning rate of $2\times10^{-5}$, a OneCycle learning-rate schedule and gradient-norm clipping at $1.0$. The 9B model is trained using two GPU processes, with a microbatch size of 1 per process and eight gradient-accumulation steps, giving a global effective batch size of 16. Training runs for three epochs, and the checkpoint saved at optimization step 2{,}000 is used for deployment.

At deployment, each task uses a fitted admission threshold $\tau_t$. The admission decision is
\begin{equation}
 \operatorname{Admit}(s_i,t_i)
 = \mathbf{1}\!\left[p_\theta(1\mid s_i)\geq\tau_{t_i}\right].
 \label{eq:jev_admission}
\end{equation}
Admitted per-case lessons enter the library with an initial utility of $0.5$. Rejected per-case lessons are retained with negative utility and excluded from retrieval-based injection during testing. Candidates rejected during consolidation are not added to L3; their review records are retained. Admission decisions are performed locally by JevGate.

\subsection*{Tools and admission control}

HealthJev uses tools to obtain task evidence. Tools return structured information, calibration notes or risk flags, and the agent determines when to call them and how to use the returned evidence. The final answer or prediction is produced by the agent.

All tasks share three built-in tools: semantic memory recall over the five-layer memory, experience writeback and external web retrieval. During testing, the experience library is frozen and writeback is disabled. Each biomedical task also uses tools matched to its evidence type: diabetes readmission uses biomedical literature search and a statistical readmission-prior lookup; GeneTuring uses a unified gene-database entry point that routes queries to MyGene, Ensembl, HGNC, NCBI E-utilities and Open Targets; MedCalc uses a structured calculator covering the 55 reference formulas; and MIMIC-ICD uses a dictionary lookup over candidate ICD-10 codes. ICU SOFA and ODIR5K use no task-specific tools, relying on memory retrieval and direct multimodal interpretation, respectively.

In addition to the tools callable by the agent, the architecture contains a non-generative decision component. Before a candidate lesson is admitted to the reusable library, JevGate, built on Kev-9B, determines admission locally. This component does not generate free text; its training is described above.

Tool inputs are restricted to information visible for the current case. Reference labels, reference answers, future cases and source-identifying paths are excluded. Reference answers are used for review and admission screening only after the case answer has been submitted and scored. The base condition receives no tools or memory and answers the same test cases directly. All conditions use the same cases and scoring rules. Comparisons with the base condition evaluate the combined tool- and memory-augmented workflow, whereas comparisons on the same scaffold assess differences between memory mechanisms.

\subsection*{Thirty-day dietary-management case}

The dietary-management case followed a simulated user with prediabetes for more than 30 days (Fig.~\ref{fig:architecture}d). It examined changes in the memory state used for subsequent recommendations; glycaemic outcomes and adherence were not evaluated.

At each interaction, the user requested meal guidance. HealthJev retrieved relevant profile facts, recent indicators, prior food preferences and any existing dietary-planning procedure. It generated a recommendation and applied induction after the episode. On day 1, recommendations were based primarily on general prediabetes dietary guidance; early preferences were added to L2, a candidate dinner-planning procedure was initialized in L3 and the episode was recorded in L4. By day 7, the L3 dinner procedure was revised in response to recurrent late dinners and mild post-meal feedback. By day 14, weekday and weekend patterns began to differ, and the procedure acquired context-specific branches. By day 30 and later, the system generated a monthly plan from the updated profile, revised procedure and accumulated episode traces.

\subsection*{Longitudinal benchmark construction}

The benchmark is built from MIMIC-IV. We selected 20 real patients from a screened pool of 8{,}097 candidates, with distinct chronic-disease combinations ranging from hypertension alone to five coexisting conditions. Diagnoses, medications, laboratory results and allergy records were extracted into a day-level fact history containing 19{,}468 entries. Each entry is annotated with its category, whether it is later superseded by a newer value, and its source table and field; critical values are recorded in both common units. The fact history is not supplied to any evaluated condition and is used only to generate daily profiles and score dialogues.

Each patient is paired with a clinically consistent simulated user profile (Fig.~\ref{fig:benchmark}b). Demographic attributes come from the source record, whereas occupation, household and caregiving context and expressive style are generated under diversity constraints. The profiles span ages 22--74 years and include 20 distinct occupation labels. Eight persona categories are represented: dutiful, steady, laid-back and self-experimenter (three users each), and anxious, terse, chatty and avoidant (two users each). These categories specify simulated interaction styles, not measured psychological traits or clinical diagnoses. Profiles also specify habitual phrases, dietary preferences, exercise and sleep routines, and family-care responsibilities. The same profile is used throughout each user's 90-day simulation, allowing these personal contexts to recur alongside changes in clinical information.

During interaction, the user simulator reports events, self-measured values and requests in colloquial first-person messages. Disclosure statements containing key values are generated in advance during benchmark construction and serve as the factual basis for scoring. Every clinical fact in the simulator's messages must come from the fact history, the goals of the day or previously disclosed values and is checked for numerical consistency. The simulator maintains a goal state machine that ends the day once all goals are satisfied. The number of conversational turns per day is capped.

The 1{,}800 interaction days contain 4{,}680 goals across the seven types reported above. Scripts include three kinds of assessment. First, longitudinal recall probes: 1{,}626 goals carry 1{,}696 cross-references to information disclosed 1 to 25 days earlier and cover 85.8\% of interaction days. About two thirds involve cued recall, in which the day's message mentions an earlier fact without repeating its value; the remainder require recall of information not mentioned in the day's message. Second, temporal-validity probes: 180 goals test use of the most recent value, for example asking for the current dose on day 45 after a medication change on day 12. Third, privacy-boundary scenarios: 720 interaction days, representing 40\% of all days, include third-party requests or similar situations to assess authorized disclosure and avoidance of over-disclosure. Governance probes are also registered per patient and day with reference behaviours covering four requirements: retaining key medical information, not writing sensitive everyday details to long-term memory, correcting misreported values and restricting third-party disclosures to authorized information. A total of 189 probes are registered, each audited on one or two pre-specified days, yielding 294 governance checks.

Benchmark construction uses three-way information isolation. The user simulator receives only the goal texts, completion conditions and disclosure statements. The evaluated agent receives only messages actually uttered by the user; the simulator's internal state is not included in the dialogue. Expected behaviours and governance references are available only for scoring. After construction, the benchmark underwent an information-leakage audit and was frozen; it has not been modified since.

After each interaction day, the automated judge GLM-5.2 assigns a single score from 0 to 1. Judging is blind: the prompt does not identify the evaluated system, and all five conditions share the same judging prompt. The judge receives the day's goals and reference behaviours, a day-by-day record of previously disclosed information, the full dialogue of the day and programmatic check results. Its rationale must quote the dialogue verbatim. The rubric covers six dimensions: long-term recall, temporal validity of previously reported values, numerical accuracy including correction of misreported values, response to critical readings, privacy boundaries over what is stored and disclosed, and overall fulfilment of the user's needs. Scoring combines verification against the day's values, the historical fact history and the user's own words to distinguish legitimate references to earlier disclosures from unsupported content.

HealthJev, a memory-free ReAct agent and the three memory-augmented frameworks MemoryBank, Mem0 and Memp were evaluated using Doubao-Seed-2.0-pro as the common backbone under the same interaction protocol, each over all 1{,}800 interaction days. No condition receives an initial patient profile; patient information can reach the system only through disclosure during interaction.

\subsection*{Six-task accumulation and testing}

We evaluate experience accumulation and transfer on six public biomedical tasks. Diabetes readmission prediction builds on the UCI Diabetes 130-US hospitals dataset \cite{clore2014diabetes}: the task is three-class readmission prediction, with 500 accumulation cases and 200 test cases. ICU SOFA estimation builds on the PhysioNet/Computing in Cardiology Challenge 2012 dataset \cite{silva2012physionet,goldberger2000physionet}, with 500 accumulation cases and 200 test cases. Exact agreement with the reference total SOFA score receives 1 point, an absolute difference of 1 receives 0.5 points, and other predictions receive 0 points. Discharge-summary coding builds on MIMIC-IV \cite{johnson2023mimiciv}: candidate codes are restricted to the 30 most frequent ICD-10 codes, predictions are scored by set F1 against the reference codes, and the split contains 502 accumulation cases and 201 test cases. Medical calculation builds on MedCalc-Bench \cite{khandekar2024medcalc} and is scored by matching values, dates and text, with 382 accumulation cases and 200 test cases. Fundus-image diagnosis builds on ODIR-5K \cite{odir2019} and is formulated as eight-class diagnosis, with 500 accumulation cases and 203 test cases. Genomic question answering builds on GeneTuring \cite{shang2025geneturing}, with 300 accumulation cases and 150 test cases. Numeric answers are scored by numerical agreement and short textual answers by normalized reference-answer matching; long sequence answers receive partial credit based on the normalized sequence similarity computed with SequenceMatcher. All conditions use Doubao-Seed-2.0-pro as the backbone model.

On each task, the accumulation split is disjoint from the test set. The agent answers cases in sequence, using multi-step reasoning and task tools as needed, after which programmatic scoring assigns a score and an error attribution. After each case, the system distils one candidate lesson: incorrect cases yield lessons describing the problem to be corrected, whereas correct cases yield lessons describing effective methods. JevGate screens each candidate. Admitted lessons enter the task's reusable experience library, and rejected lessons are excluded from subsequent test-time retrieval. In addition to per-case induction, a consolidation step every 20 cases aggregates recent episodes into additional candidate lessons, which are also screened by JevGate. After accumulation, each task's library is frozen and exported for within-task testing and cross-task transfer evaluation.

Accumulation and test splits are separated at the patient or subject level where applicable and audited for overlap. All conditions use the same test cases and evaluation protocol. The base condition answers with the backbone alone, without memory or tools. HealthJev uses the frozen task library and injects, read-only, the six most relevant lessons retrieved by semantic search. ReAct, MemoryBank, Mem0 and Memp run their respective mechanisms on the same scaffold, with the same test cases and scoring. Comparisons with the base condition include the combined contribution of tools and the experience library, whereas comparisons on the same scaffold assess differences between memory mechanisms.

Scoring is deterministic and programmatic on all tasks: classification tasks use exact match, ICD coding uses set F1, and medical calculation uses matching of values, dates and text. GeneTuring and ICU SOFA use the task-specific partial-credit rules described above. Task metrics are the mean case scores. No model-based judgement is used. Binarization for significance testing is described in the statistical analysis.

\subsection*{Cross-backbone and cross-task transfer}

To evaluate cross-backbone transfer, we mounted the six task libraries accumulated with Doubao-Seed-2.0-pro on six further backbones: DeepSeek-V4.1-Flash, GPT-5.4, GPT-4.1, Gemini-3.8-Flash, Qwen3.5-Plus and a self-hosted Qwen3.8-27B. The libraries remain frozen and read-only during transfer. Retrieval and injection follow the within-task testing protocol, and lessons come only from accumulation on the original backbone. The reference for each backbone is its own base condition, which answers the same test cases directly without memory or tools; scoring is identical to the main evaluation. The libraries improved the six-task macro-average on all seven backbones, including the original backbone. Per-task significance is assessed with exact McNemar tests and macro-level significance with a two-sided exact Wilcoxon test over the six tasks.

Cross-task transfer is evaluated on the accumulation backbone. The frozen library of task A is mounted on the test cases of task B. Lessons come only from the accumulation phase of task A and do not overlap with the test cases of task B; retrieval and injection are unchanged. The reference is the base condition of task B. Transfer across eight donor--target pairs yielded gains of 0.3 to 29.5 percentage points over the base condition, with the largest gains from transferring the GeneTuring and ICU SOFA libraries to diabetes readmission. Both gains were significant by a two-sided exact McNemar test, with $p<0.05$.

\subsection*{Statistical analysis}

For the longitudinal benchmark, the unit of pairing is the patient--day interaction. Comparisons between HealthJev and each reference condition use two-sided paired Wilcoxon signed-rank tests. The 95\% confidence intervals of the per-condition mean daily scores are obtained from a user-level cluster bootstrap that resamples the 20 patients as whole clusters, with 10{,}000 resamples and random seed 20260705. At the user level, we report the number of patients whose mean daily score is higher under HealthJev than under the strongest reference condition.

For the six biomedical tasks, the unit of pairing is the test case. Each condition is compared with the base condition using two-sided exact McNemar tests. Case-level outcomes are binarized at a positive score: a case counts as a success when its score is greater than zero. Benjamini--Hochberg false-discovery-rate correction is applied within each family of condition-versus-base comparisons. The $p$ values reported in the text are uncorrected, and the pattern of significance is unchanged after correction.

For cross-backbone transfer, each backbone-by-task combination is assessed with an exact McNemar test comparing the mounted library with the same backbone's base condition, using the same binarization. Macro-average significance is assessed with a two-sided exact Wilcoxon test over the six tasks. The eight donor--target pairs of cross-task transfer are assessed with exact McNemar tests. All tests are two-sided.

\subsection*{Ethics and data governance}

The MIMIC-IV data used in this study were obtained under PhysioNet credentialed access and data use agreements, with the required training and approval completed. The longitudinal benchmark is built solely from de-identified records: patient identifiers are removed, and source records enter the fact history only as day-level clinical facts, including diagnoses, medications, laboratory categories and values. Dialogues are simulated first-person narratives and contain no re-identifiable text from the source records.

Benchmark scripts were audited for information leakage and frozen. The released scripts are synthetic narratives derived from de-identified records and contain neither patient identifiers nor source-record text. As controlled products derived from MIMIC, the complete fact history and patient intake profiles are not released directly. Researchers with the required access can regenerate them from MIMIC-IV under its access terms by following the released reproduction pipeline.

The six biomedical tasks use public or controlled public datasets and public tool resources, obtained and used under the original providers' terms. We do not redistribute raw medical images, clinical records or third-party model checkpoints.

All evaluations were performed on offline benchmarks; patient outcomes, real-world safety and clinical effectiveness were not evaluated. HealthJev is positioned here as an assistive system for longitudinal personal health management and information organization. Deployment for diagnosis, treatment recommendation or autonomous intervention would require separate ethical, regulatory and security review and prospective clinical evaluation.

\section*{Data availability}

The frozen version of the longitudinal benchmark constructed in this
study---including the 20-patient, 90-day interaction scripts, the goal and
governance-probe annotations and the evaluation protocol---is released with
the repository below; the released scripts are synthetic narratives derived
from de-identified records and contain neither patient identifiers nor text
from the source records.

The six biomedical tasks use public datasets: the UCI Diabetes 130-US
hospitals dataset \cite{clore2014diabetes}, GeneTuring
\cite{shang2025geneturing}, ODIR-5K \cite{odir2019}, the PhysioNet/Computing in
Cardiology Challenge 2012 dataset \cite{silva2012physionet,goldberger2000physionet}, MIMIC-IV
\cite{johnson2023mimiciv} and MedCalc-Bench \cite{khandekar2024medcalc}. These
datasets should be obtained from their original providers under their access
terms; the authors do not redistribute raw images, clinical records, the fact
history or third-party model checkpoints.

\section*{Code availability}

HealthJev is publicly available at \url{https://github.com/HC-Guo/Self-Evolving_HealthJev}.
The repository contains the implementation of the agent framework and the
five-layer memory, the training and calibration code of the JevGate admission
model, the construction scripts and frozen assets of the longitudinal
benchmark, and the evaluation and scoring scripts for the six biomedical tasks.

\section*{Acknowledgements}

This work was supported by Lingang Laboratory (No. LGL-1987-11, LGL-2930-01),
the 111 Project (No. B18015) and the Yangtze River Delta Science and
Technology Innovation Community Joint Research (Basic Research) Project
(No. 2025CSJZN00100).

\section*{Author contributions}

H.L. and J.D. contributed equally to this work. H.L., J.D., Z.W. and H.G.
developed the core codebase and system implementation. H.L., J.H. and Y.W.
performed the benchmark experiments and evaluation analyses. H.L., J.D. and
H.G. drafted the manuscript. H.G. conceived and led the project and
supervised the study. J.F. and X.Z. provided senior guidance, domain
expertise and critical manuscript revision. T.J., J.S., W.G., W.C., J.Y. and Q.S.
contributed domain expertise, resources and manuscript revision. All authors
discussed the results, reviewed the manuscript and approved the final
version.

\section*{Competing interests}

Q.S. is an employee of JD.com, Inc. The other authors declare no competing
interests.

\end{document}